\pdfoutput=1

\documentclass[11pt]{article}

\usepackage{latex/acl}
\usepackage{times}
\usepackage{latexsym}

\usepackage[T1]{fontenc}

\usepackage[utf8]{inputenc}

\usepackage{microtype}

\usepackage{inconsolata}

\usepackage{amsfonts,amssymb}
\usepackage{amsmath}
\usepackage{multirow}
\usepackage{colortbl}
\usepackage{booktabs}
\usepackage{graphicx}
\usepackage{hyperref}

\title{Fusion Makes Perfection: An Efficient Multi-Grained Matching Approach for Zero-Shot Relation Extraction}

\author{Shilong Li\thanks{The first three authors contribute equally.}, Ge Bai\footnotemark[1], Zhang Zhang\footnotemark[1], Ying Liu \\  {\bf Chenji Lu}, {\bf Daichi Guo}, {\bf Ruifang Liu}, {\bf Yong Sun}\thanks{Yong Sun is the corresponding author.} \\
        Beijing University of Posts and Telecommunications, China\\
        \texttt{lishilong2019210645@bupt.edu.cn}}

\begin{document}
\maketitle
\begin{abstract}
Predicting unseen relations that cannot be observed during the training phase is a challenging task in relation extraction. 
Previous works have made progress by matching the semantics between input instances and label descriptions. However, fine-grained matching often requires laborious manual annotation, and rich interactions between instances and label descriptions come with significant computational overhead. 
In this work, we propose an efficient multi-grained matching approach that uses virtual entity matching to reduce manual annotation cost, and fuses coarse-grained recall and fine-grained classification for rich interactions with guaranteed inference speed.
Experimental results show that our approach outperforms the previous State Of The Art (SOTA) methods, and achieves a balance between inference efficiency and prediction accuracy in zero-shot relation extraction tasks.
Our code is available at \href{https://github.com/longls777/EMMA}{https://github.com/longls777/EMMA}.
\end{abstract}

\section{Introduction}
Relation Extraction (RE) is an important task of Natural Language Processing (NLP), which aims to identify the relation between a pair of entities within a given sentence.
Previous RE models \citep{han-etal-2020-data, peng2020learning, zhao-etal-2021-relation} have impressive performance through large-scale supervised learning based on high-quality labeled data. However, collecting sufficient data for every new relation type is laborious in practice.
This leads to the necessity of zero-shot RE task, which involves extracting unobserved relations.

Recently, semantic matching \citep{obamuyide-vlachos-2018-zero} has become a mainstream paradigm of zero-shot RE, which matches a given input with a corresponding label description.
PromptMatch \citep{sainz-etal-2021-label} performed self-attention over each instance-description pair to enrich interaction, but increased computational overhead.
ZS-Bert \citep{chen-li-2021-zs} enabled fast inference by encoding the input and description separately, and then storing and reusing the representation of descriptions for each input. However, the lack of interaction during the encoding also limits the performance of the model. 
RE-Matching \citep{zhao-etal-2023-matching} introduced a unique fine-grained matching pattern and improved both the accuracy and speed by ignoring redundant components in the instance and matching the entities with their hypernyms in the description.
However, this approach relies on manual annotation of entity hypernyms in label descriptions and still lacks the interaction between instances and descriptions.
Therefore, how to achieve a balance between efficiency and accuracy without using additional labor costs is a pressing issue.
\begin{figure}[!t]
\setlength{\abovecaptionskip}{0.1cm}  
\setlength{\belowcaptionskip}{-0.6cm}
    \centering
    \includegraphics[width=0.4\textwidth,height=0.35\textwidth]{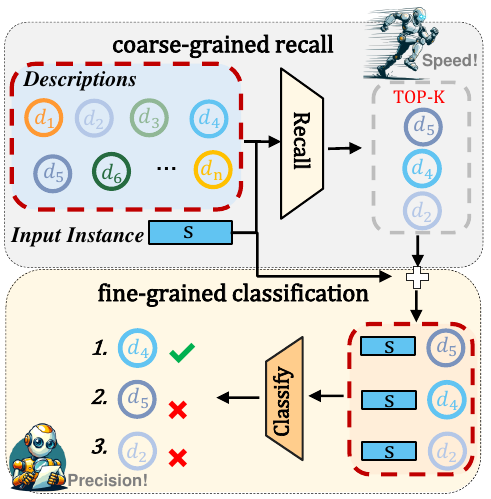}
    \caption{The overall process of our method. The coarse-grained recall refers to the rough and rapid screening of several possible results, while the fine-grained classification denotes the detailed discrimination of these possible results.}
    \label{intro}
\end{figure}

\begin{figure*}[!t]
\centering
\includegraphics[width=2.1\columnwidth,height=0.9\columnwidth]{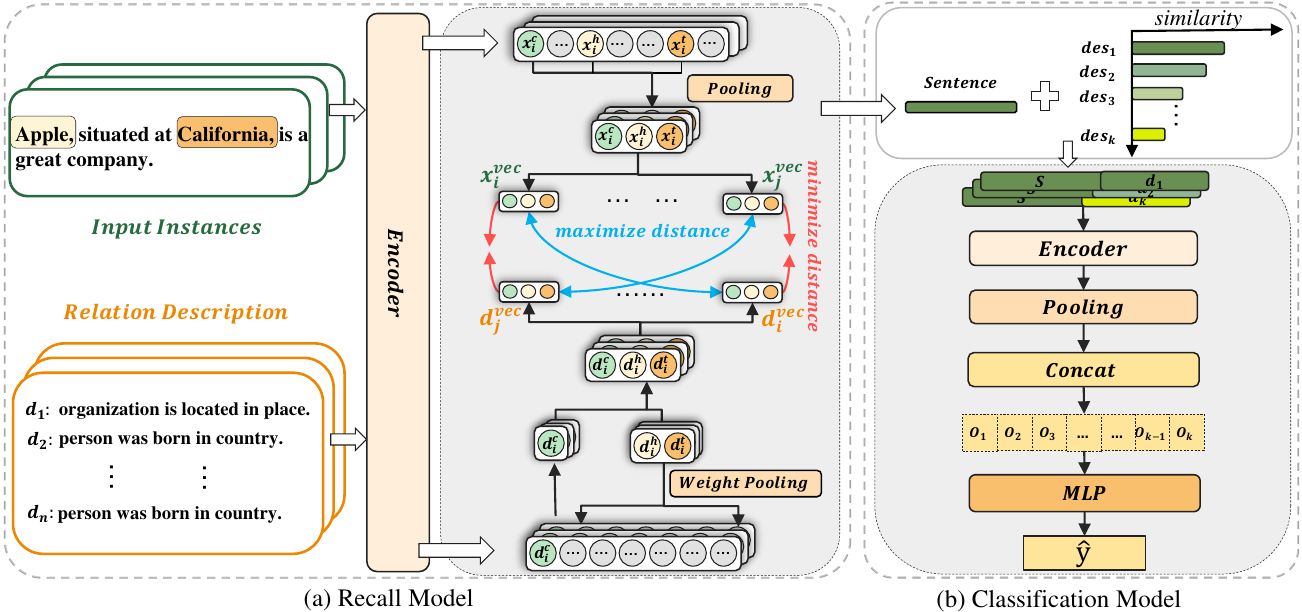}
\caption{The overall architecture of EMMA. 
(a) The recall model swiftly matches to retrieve the top $k$ most probable relations.
(b) The classification model further distinguishes among these similar relations.}
\label{model}
\end{figure*}

To address this issue, we propose an \textbf{E}fficient \textbf{M}ulti-Grained \textbf{M}atching \textbf{A}pproach (EMMA). 
In this work, we generate virtual entity representations of descriptions in semantic matching instead of annotating descriptions to avoid manual costs. Additionally, we utilize a fusion of coarse-grained recall and fine-grained classification. Specifically, a coarse-grained filter is used to improve inference speed and select several candidate relations for each input, while a fine-grained classifier enhances instance-description interaction, enabling more accurate selection from relation candidates to improve prediction precision.

We summarize the contributions as follows:
\begin{itemize}
    \item To the best of our knowledge, EMMA is the first work fusing the coarse-grained recall stage and fine-grained classification stage to achieve a balance of accuracy and inference speed.
    \item We introduce a virtual entity matching method to achieve effective semantic matching as well as avoid laborious manual annotation.
    \item Extensive experiments on different datasets and settings show EMMA outperforms previous SOTA methods, which demonstrates the efficiency and effectiveness of our approach. 
\end{itemize}

\section{Approach}
\subsection{Task Formulation}
The zero-shot RE task is designed to learn from the seen relations $\mathcal{R}_s=\{r_1^s,r_2^s,...,r_n^s \}$ to identify unseen relations $\mathcal{R}_u=\{r_1^u,r_2^u,...,r_m^u \}$. These two sets are disjoint, and the model only uses $\mathcal{R}_s$ during the training phase. Similar to the previous work \citep{zhao-etal-2023-matching, chen-li-2021-zs}, we formulate zero-shot RE as a semantic matching task. We further subdivide it into two stages: recall stage and classification stage.

In the recall stage, the training set comprises $N$ samples $\mathcal{D}=\{(x_i,e_i^h, e_i^t, y_i,d_i)|i=1,...,N\}$, where $x_i$ is input instance,  $e_i^h$ is head entity, $e_i^t$ is tail entity, $y_i \in \mathcal{R}_s$ is corresponding relation and $d_i$ is the relation description. We optimize a recall model $\mathcal{M}_r(x,e^h,e^t,d) \to s \in \mathbb{R}$ on $\mathcal{R}_s$, where $s$ represents the matching score between the instance and description. Then we recall top $k$ relation exhibiting the highest matching scores.

In the classification stage, for the instance and the top $k$ relation descriptions, we optimize a fine-grained classification model $\mathcal{M}_c(x,e^h,e^t,d_1,d_2,...,d_k) \to \hat{y}$, where $\hat{y}$ is the predicted probability. 


During testing on $\mathcal{R}_u$, given a sample $(x_u, e_u^h,e_u^t)$, we use $\mathcal{M}_r$ to obtain the top $k$ most probable relation at a coarse-grained level, and use $\mathcal{M}_c$ to further distinguish these relations at a fine-grained level, obtaining the most probable one.

\subsection{Coarse-grained recall}
To rapidly query the relation corresponding to the input instance without tediously encoding and matching each pair \citep{sainz-etal-2021-label}, we adopt a dual-tower-like architecture \citep{Yi2019SamplingbiascorrectedNM}, which allows for precomputing representations of numerous relations to facilitates swift matching.
\subsubsection{Input Instance Encoder}
\label{221}
Given an input instance $x=\{w^x_1,...,w^x_n\}$, distinct special tokens $[E_h]$, $[\textbackslash E_h]$, $[E_t]$, $[\textbackslash E_t]$ are employed to wrap the head entity and tail entity, respectively. 
After inputting $x$ into a pre-trained encoder, we utilize the last hidden states of special tokens $[E_h]$, $[E_t]$, and $[CLS]$ (refer to $w_0^x$) as representations of head entity, tail entity, and contextual information, which is formulated as follows:
\begin{equation}
\mathit{h}^x_0,\mathit{h}^x_1,...,\mathit{h}^x_n = \mathbf{BERT}(w^x_0,w^x_1,...,w^x_n)\label{e1}
\end{equation}
\begin{equation}
    \mathit{x}^c = \mathit{h}^x_0, \mathit{x}^h = \mathit{h}^x_{[E_h]},\mathit{x}^t = \mathit{h}^x_{[E_t]}
\end{equation}
Then we combine the representation of head entity $x^h \in \mathbb{R}^d$, tail entity $x^t \in \mathbb{R}^d$, and the contextual information $x^c \in \mathbb{R}^d$ to form the comprehensive representation $x^{vec} \in \mathbb{R}^{3d}$ of the input instance.
\begin{equation}
x^{vec} = \mathit{x}^c \oplus \mathit{x}^h \oplus \mathit{x}^t
\end{equation}
where $d$ is the hidden dimension of the encoder and $\oplus$ denotes the concatenation operator.
\subsubsection{Virtual Entity Matching}
Although the description of corresponding relation $d=\{w^d_1,...,w^d_n\}$ is easily obtainable (e.g. from Wikipedia), manually annotating the entity hypernyms within various relations is still time-consuming and laborious. 
Therefore, we directly input relation descriptions into the pre-trained encoder. Then, we employ two weight pooling layers with different parameters to obtain separate virtual entity representations $d^h \in \mathbb{R}^{d}$ and $d^t \in \mathbb{R}^{d}$. Similar to Section \ref{221}, we use the hidden states corresponding to the $[CLS]$ token (refer to $w^d_0$) as the contextual representation $d^c \in \mathbb{R}^{d}$, and concatenate these three to obtain the comprehensive representation $d^{vec} \in \mathbb{R}^{3d}$ of the relation description.
\begin{equation}
\mathit{h}^d_0,\mathit{h}^d_1,...,\mathit{h}^d_n = \mathbf{BERT}(w^d_0,w^d_1,...,w^d_n)\label{e1}
\end{equation}
\begin{equation}
    \mathit{d}^c = \mathit{h}^d_0
\end{equation}
\begin{equation}
    \mathit{d}^h = \textit{WeightPooling}_1(\mathit{h}^d_1,...,\mathit{h}^d_n)
\end{equation}
\begin{equation}
    \mathit{d}^t = \textit{WeightPooling}_2(\mathit{h}^d_1,...,\mathit{h}^d_n)
\end{equation}
\begin{equation}
    d^{vec} = \mathit{d}^c \oplus \mathit{d}^h \oplus \mathit{d}^t
\end{equation}
For the weight pooling, we employ the scheme proposed by \citet{Lin2017ASS}, utilizing an attention mechanism over the last hidden states of the pre-trained encoder to generate representations of virtual entities, which is formulated as follows:
\begin{equation}
    H = (h^d_1,...,h^d_n)
    \vspace{-0.1cm}
\end{equation}
\begin{equation}
    A = \mathit{softmax}(HW+b)
\end{equation}
\begin{equation}
    d^* = AH
\end{equation}
where $H \in \mathbb{R}^{(L-1)\times d}$ is the last hidden states of the encoder excluding $[CLS]$ token ($L$ denotes the max sequence length). $W$ is a linear layer of $(L-1) \times 1$, $b \in \mathbb{R}^{L-1}$ is the bias, and $A \in \mathbb{R}^{L-1}$ denotes the final weights. The final representation $d^* \in \mathbb{R}^d$ is obtained by weighting $H$ using $A$.
\subsubsection{Contrastive Learning}
When $N$ input instances $\{x_1,...,x_N\}$ and their corresponding relation descriptions $\{d_1,...,d_N\}$ are input into the encoder within a mini-batch, we obtain the representations of instance $x^{vec}_i$ and description $d^{vec}_i$, $i \in [1, N]$. To effectively learn the matching relationship between $x^{vec}_i$ and $d^{vec}_i$, we utilize a contrastive learning method, where $d^{vec}_i$ serves as a positive sample and other $N-1$ samples within the mini-batch $d^{vec}_j(j\neq i)$ serve as negative samples. The goal of contrastive learning is to minimize the distance between $x^{vec}_i$ and $d^{vec}_i$ while maximizing the distance from $d^{vec}_j$.

We utilize cosine similarity as the measurement and employ the infoNCE\citep{Oord2018RepresentationLW} as the contrastive loss function:
\begin{equation}
\mathcal{L}_i = -\log \frac{e^{\text{sim}(x^{vec}_i, d^{vec}_i) / \tau}}{\sum_{j=1}^{N} e^{\text{sim}(x^{vec}_i, d^{vec}_j) / \tau}}
\end{equation}
where $\tau$ is a temperature hyperparameter and $\text{sim}(\cdot)$ is the cosine similarity.
\definecolor{mygray}{gray}{.9}
\begin{table*}[!t]
\centering
\resizebox{\textwidth}{!}{
    \begin{tabular}{clcccccc}
    \bottomrule
    \multirow{2}{*}{\textbf{Unseen Labels}} & \multirow{2}{*}{\textbf{Method}} & \multicolumn{3}{c}{Wiki-ZSL} & \multicolumn{3}{c}{FewRel} \\ \cline{3-8} 
     & & Prec. & Rec. & $F_1$ & Prec. & Rec. & $F_1$ \\ \hline
    \multirow{6}{*}{m=5} & ZS-BERT\citep{chen-li-2021-zs} & 71.54 & 72.39 & 71.96 & 76.96 & 78.86 & 77.90 \\
                        & PromptMatch\citep{sainz-etal-2021-label} & 77.39 & 75.90 & 76.63 & 91.14 & 90.86 & 91.00 \\
                         & REPrompt\citep{chia-etal-2022-relationprompt} & 70.66 & 83.75 & 76.63 & 90.15 & 88.50 & 89.30 \\
                         & RE-Matching\citep{zhao-etal-2023-matching} & 78.19 & 78.41 & 78.30 & 92.82 & 92.34 & 92.58 \\ 
                         &  EMMA(onlyRecall) & 89.30 & 90.10 & 89.70 & 93.68 & 92.76 & 93.22 \\  
                         &  \textbf{EMMA} & \textbf{91.32} & \textbf{90.65} & \textbf{90.98} & \textbf{94.87} & \textbf{94.48} & \textbf{94.67} \\  \hline\hline
    \multirow{6}{*}{m=10}    & ZS-BERT\citep{chen-li-2021-zs} & 60.51 & 60.98 & 60.74 & 56.92 & 57.59 & 57.25 \\
                         & PromptMatch\citep{sainz-etal-2021-label} & 71.86 & 71.14 & 71.50 & 83.05 & 82.55 & 82.80 \\
                         & REPrompt\citep{chia-etal-2022-relationprompt} & 68.51 & 74.76 & 71.50 & 80.33 & 79.62 & 79.96 \\
                         & RE-Matching\citep{zhao-etal-2023-matching} & 74.39 & 73.54 & 73.96 & 83.21 & 82.64 & 82.93 \\ 
                         &  EMMA(onlyRecall) & 85.99 & 84.37 & 85.17 & 86.67 & 84.32 & 85.48 \\ 
                         &   EMMA & \textbf{86.00} & \textbf{84.55} & \textbf{85.27} & \textbf{87.97} & \textbf{86.48} & \textbf{87.22} \\  \hline\hline
    \multirow{6}{*}{m=15}    & ZS-BERT\citep{chen-li-2021-zs} & 34.12 & 34.38 & 34.25 & 35.54 & 38.19 & 36.82 \\
                         & PromptMatch\citep{sainz-etal-2021-label} & 62.13 & 61.76 & 61.95 & 72.83 & 72.10 & 72.46 \\
                         & REPrompt\citep{chia-etal-2022-relationprompt} & 63.69 & 67.93 & 65.74 & 74.33 & 72.51 & 73.40 \\
                         & RE-Matching\citep{zhao-etal-2023-matching} & 67.31 & 67.33 & 67.32 & 73.80 & 73.52 & 73.66 \\ 
                         &  EMMA(onlyRecall) & 76.83 & 75.79 & 76.31 & 78.24 & 75.77 & 76.99 \\  
                         &  \textbf{EMMA} & \textbf{78.51} & \textbf{77.63} & \textbf{78.07} & \textbf{80.47} & \textbf{79.73} & \textbf{80.10}  \\ \bottomrule
    \end{tabular}
}
\caption{Main results on Wiki-ZSL and FewRel dataset. We report the average results obtained from running with five random seeds ($k=2$) and the improvement is significant (using a Wilcoxon signed-rank test; p < 0.05). }
\label{mainResult}
\end{table*}

\subsection{Fine-grained classification}
In the recall stage, 
we obtain representations of input instances and relation descriptions separately for quick query matching.
However, the lack of interaction between the instances and descriptions limits the model's performance ceiling.
To tackle this issue, we propose fine-grained classification after coarse-grained recall, which jointly encodes instances and descriptions.

In the classification stage, during training, for each input instance $x$, $k$ relation descriptions $\mathit{D}=\{d_1,...,d_k\}$ are selected from the mini-batch of the recall stage, which includes $d_+$ corresponding to the entity relation of $x$, and top $k-1$ descriptions with the highest matching scores excluding $d_+$. The objective of classification is to select $d_+$ from $\mathit{D}$. We formulate this process as follows:
\begin{equation}
    O_i = \mathit{Pooling}(BERT(\langle x \oplus d_j \rangle))
\end{equation}
\begin{equation}
    \hat{y}=\mathit{MLP}(O_0 \oplus O_1 \oplus ... O_k)
\end{equation}
\begin{equation}
    \mathcal{L}_c=-log(\frac{e^{\hat{y}_+}}{\sum_{i=1}^{k}{e^{\hat{y}_i}}})
\end{equation}
where $O_i$ is the representation of instance-description pair obtained by extracting the last hidden state of the $[CLS]$ token and $\hat{y}$ is the predicted probability. We utilize cross-entropy as the loss function for classification.


During testing, the top $k$ descriptions with the highest matching scores are selected as input.

\section{Experiments}
We conduct our experiments on the FewRel \citep{han-etal-2018-fewrel} and Wiki-ZSL \citep{chen-li-2021-zs} datasets. Specific details about the datasets and experimental details are provided in appendix \ref{setup}.
\subsection{Main results}
Table \ref{mainResult} displays the experimental results on the Wiki-ZSL and FewRel datasets, showing that our proposed method significantly outperforms the previous SOTA results by a large margin when predicting different numbers of unseen relations, specifically when $m=15$, it achieves at least a $\mathbf{11}\%$ improvement in F1 scores on Wiki-ZSL and a $\mathbf{6}\%$ improvement on FewRel. Even the EMMA model without the classification (onlyRecall), which selects the relation with the highest matching score as the prediction result, still outperforms the SOTA model. Moreover, compared to RE-Matching, EMMA employs virtual entity matching, avoiding the human effort required for annotated descriptions.
Upon integrating the classification model, the complete version of EMMA extensively augments the interaction between the input sentence and relation description, further boosting the model's performance. These showcase the superiority of our model.

\begin{table}[!t]
\centering
\begin{tabular}{clccc}
\bottomrule
                  \textbf{Dataset} & \textbf{Method}  & Prec. & Rec. & $F_1$ \\ \hline
\multirow{4}{*}{\textbf{FewRel}} & w/o Vir. & 76.43 & 76.02 & 76.22 \\
                  & w/o Cla. & 78.24 & 75.77 & 76.99 \\
                  & w/o both & 75.54 & 75.12 & 75.33 \\
                  & \textbf{Ours} & \textbf{80.47} & \textbf{79.73} & \textbf{80.10} \\ \bottomrule
\end{tabular}
\caption{Ablation study on FewRel ($m=15$, $k=2$).}
\label{ablation}
\end{table}

\subsection{Ablation Study}
Table \ref{ablation} presents the results of ablation experiments, which indicates that removing the virtual entity matching (w/o Virt.) and the classification (w/o Cla.) individually both result in decreased model performance. 
This illustrates the effectiveness of virtual entity matching during the recall stage and the efficacy of the classification model designed to enhance interaction for identifying similar relations.
When both are removed (w/o both), the model degrades to a simple semantic matching model, leading to a significant decline in performance.
\begin{figure}[!t]
\centering
\includegraphics[width=1\linewidth]{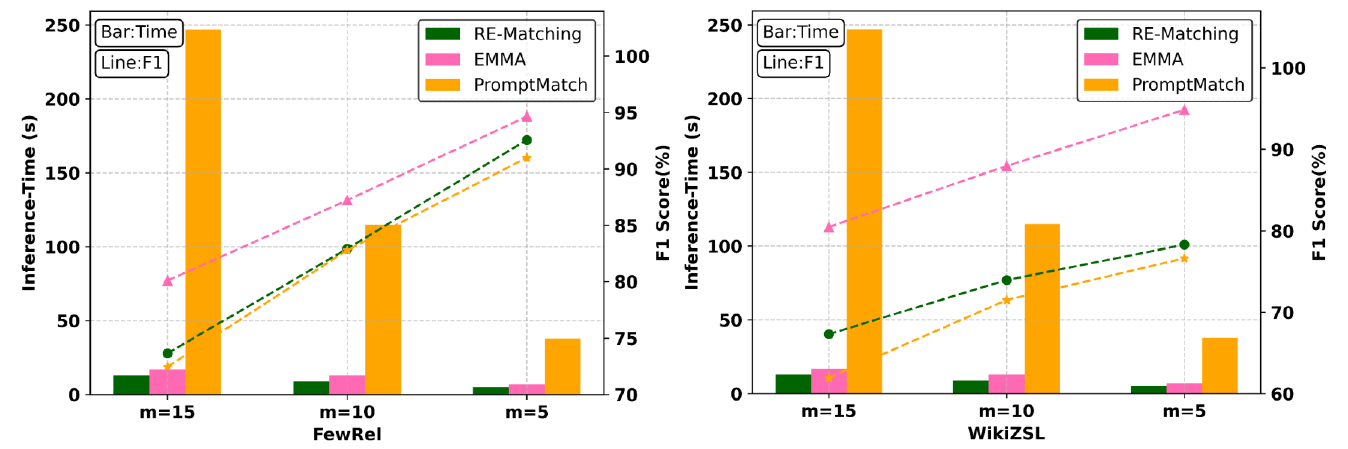}
\caption{Comparison in terms of runtime(Bars) and matching F1 (Dotted lines).}
\label{compare}
\end{figure}

\subsection{Inference Efficiency}
Figure \ref{compare} shows the inference runtime and matching F1 scores. As the number of new relations $m$ increases, EMMA proves more efficient than PromptMatch. While it takes slightly longer than RE-Matching, EMMA significantly improves F1 scores. Detailed analysis is in appendix \ref{infer_ala}.

\section{Conclusions}
In this work, we propose a fusion method for ZeroRE named EMMA, which enhances performance in the ZeroRE task by combining coarse-grained recall and fine-grained classification, while maintaining efficient inference capabilities. 
Experimental results demonstrate that our approach outperforms SOTA methods in matching F1 scores while maintaining rapid inference.

\section{Limitations}
Our proposed method has only been experimented on zero-shot relation extraction tasks and has not been applied in other domains of information extraction, such as named entity recognition.
However, the underlying principles embedded within EMMA might potentially be generalized and applied to other related tasks.

\bibliography{custom}

\appendix
\section{Experimental setup}
\label{setup}
\subsection{Datasets}
\textbf{FewRel} \citep{han-etal-2018-fewrel} is a 
dataset designed for few-shot relation classification. It's sourced from Wikipedia and involves manual annotation by crowd workers. It comprises 80 relations, each having 700 associated sentences. \textbf{Wiki-ZSL} \citep{chen-li-2021-zs} originates from the Wikidata Knowledge Base, boasting 94,383 sentences spanning across 113 relation types. In Wiki-ZSL, entities are extracted from Wikipedia articles and linked to the Wikidata knowledge base. This method of remotely supervised generation results in Wiki-ZSL containing more noise than FewRel.

\begin{figure*}[!t]
\centering
\includegraphics[width=2.0\columnwidth,height=0.5\columnwidth]{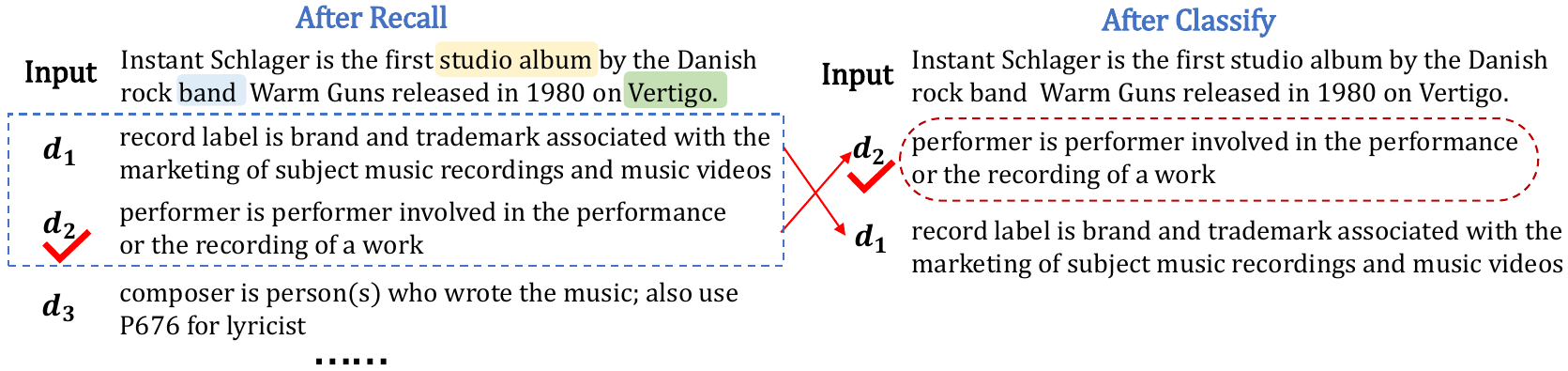}
\caption{This is an example showcasing the role of the classification model. During the recall stage, the correct relation description wasn't ranked first, yet through the fine-grained classification model's correction, the accurate result was eventually obtained.}
\label{case}
\end{figure*}
For the accuracy and comparability of experimental results, similar to \citet{zhao-etal-2023-matching}, we randomly selected $m \in \{5, 10, 15\}$ relations as the validation set, $m$ relations as the test set, and the remainder as the training set. Simultaneously, we chose $5$ different random seeds for dataset partitioning and experimentation, reporting the average results of these experiments.

\begin{figure}[!t]
\centering
\includegraphics[width=1\linewidth]{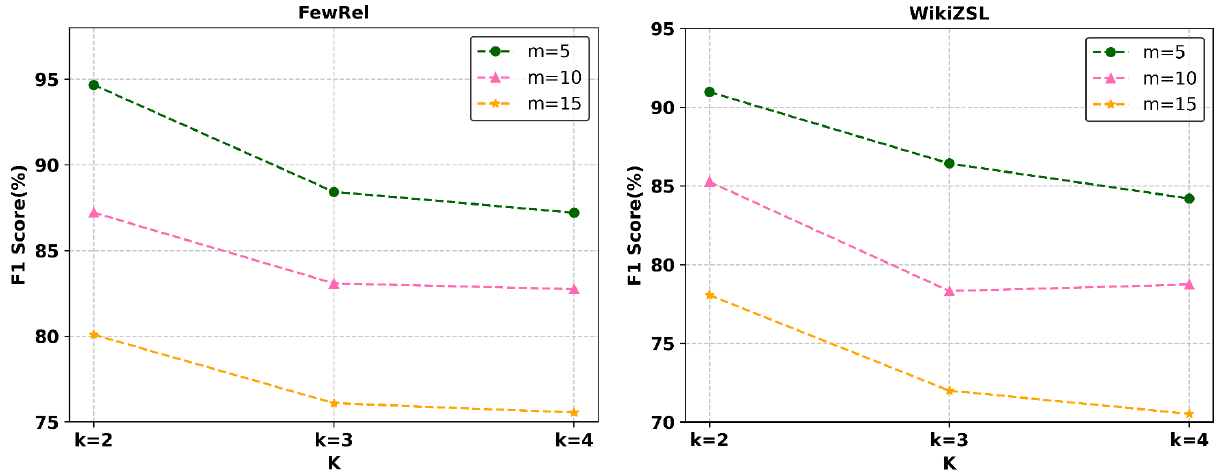}
\caption{The F1 scores of EMMA across different values of $k$.}
\label{k}
\end{figure}

\subsection{Implementation Details}
We utilize \emph{Bert-base-uncased} as the pre-trained encoder, which is then fine-tuned for our purposes. In the recall model, the encoder for the input sentence shares parameters with the encoder for the relation description. The encoder in the classification model has its separate parameters. The recall model and classification model are jointly trained in the experiment and we discuss the differences between joint training and separate training in appendix \ref{d_t_m}. 

The temperature $\tau$ for the infoNCE loss is set to 0.02. We use AdamW optimizer with a learning rate of $2e-5$ and a batch size of $64$. We train the model for $5$ epochs with a warm-up of 100 steps. All experiments are conducted using an NVIDIA RTX A6000.

\section{Ablation Experiments on Wiki-ZSL}

\begin{table}[!htbp]
\centering
\begin{tabular}{clccc}
\bottomrule
                  \textbf{Dataset} & \textbf{Method}  & Prec. & Rec. & $F_1$ \\ \hline
\multirow{4}{*}{\textbf{Wiki-ZSL}} & w/o Virt.  & 74.03 & 74.74 &  74.38  \\
                  & w/o Cla. & 76.83 & 75.79 & 76.31 \\
                  & w/o both & 71.52 & 70.93 & 71.22 \\
                  & \textbf{Ours} & \textbf{78.51} & \textbf{77.63} & \textbf{78.07} \\ \bottomrule
\end{tabular}
\caption{Ablation study on Wiki-ZSL ($m=15$, $k=2$).}
\label{ablation2}
\end{table}

The ablation experiments conducted on the Wiki-ZSL dataset align with our conclusion that both virtual entity matching and classification components contribute beneficially to improving model performance.

\section{Analysis of classification performance}
Figure \ref{case} illustrates an instance where the classification model corrects a recalled result. However, it's possible for the top $1$ result obtained by the recall model to be the correct one, yet after classification, an incorrect result is generated. Nonetheless, the experimental results in Table \ref{mainResult} comparing EMMA and EMMA (onlyRecall) indicate that the number of corrections by the classification model is greater than the number of errors corrected. This demonstrates the effectiveness of fine-grained classification.

\section{Difference over various $k$}
Figure \ref{k} illustrates the change in EMMA's F1 scores across different values of $k$ on the FewRel and Wiki-ZSL datasets. As $k$ increases (from 2 to 4), the model's F1 score gradually decreases. This could be attributed to the increased difficulty in classification as the model needs to discern among a larger set of relations when $k$ grows. How to mitigate this decline in such scenarios can be considered as a future research direction.

\section{Differences between training methods}
\label{d_t_m}

\begin{table}[htbp]
\centering
\begin{tabular}{cccc}
\bottomrule
                  \textbf{Training approach} & Prec. & Rec. & $F_1$ \\ \hline
joint training & \textbf{94.87} & \textbf{94.48} & \textbf{94.67} \\ \hline
separate training & 94.62 & 94.36 & 94.49 \\ \hline
\end{tabular}
\caption{Experimental comparison on FewRel ($m=5$, $k=2$).}
\label{ec}
\end{table}
In joint training, we train the recall model and the classification model at the same time, which means that the loss from the classification stage will backpropagate to the recall stage. In separate training, the recall model is trained first, and then the classification model is trained based on the output of the recall model. Regardless of the method, we ensure that the input to the classification model includes the correct relation.

From the experimental results, it can be observed that the difference between separate training and joint training is not significant.

\section{Inference Efficiency Analysis}
\label{infer_ala}
For both RE-Matching \citep{zhao-etal-2023-matching} and EMMA, the representation vectors of relation descriptions can be pre-inferred. When inputting an instance, its obtained vector needs to be compared with each description vector individually. Assuming there are $m$ instances and $n$ relations, both models need to process this. The inference speed of RE-Matching should be $O(m*n+n)$, while EMMA, due to the inclusion of a fine-grained classification model, operates at $O(m*n+m+n)$. However, in real-world scenarios where both $m$ and $n$ are large, the time complexity of both models tends toward $O(m*n)$, making the inference speed of EMMA and RE-Matching essentially similar. Certainly, we could use neighbor search methods like HNSW \citep{Malkov2016EfficientAR} to reduce the time complexity of one-to-one matching in the recall stage. However, that is not the focus of this work.

Taking FewRel as an example, each relation comprises 700 test input instances. RE-Matching and our EMMA encode the input sentences and descriptions separately, with encoding performed $(700 \cdot n + n)$ times and $(700 \cdot n + 700 + n)$ times, respectively. In contrast, PromptMatch requires concatenation of text pairs for input and involves encoding performer $(700\cdot n^2)$ times. 

\end{document}